\documentclass{article}

\usepackage{PRIMEarxiv}

\usepackage[utf8]{inputenc} % allow utf-8 input
\usepackage[T1]{fontenc}    % use 8-bit T1 fonts
\usepackage{hyperref}       % hyperlinks
\usepackage{url}            % simple URL typesetting
\usepackage{booktabs}       % professional-quality tables
\usepackage{amsfonts}       % blackboard math symbols
\usepackage{nicefrac}       % compact symbols for 1/2, etc.
\usepackage{microtype}      % microtypography
\usepackage{lipsum}
\usepackage{fancyhdr}       % header
\usepackage{graphicx}       % graphics
\graphicspath{{media/}}     % organize your images and other figures under media/ folder

\usepackage{subcaption}
\usepackage{amsmath}
\usepackage{amsthm}
\usepackage[english]{babel}
\newtheorem{definition}{Definition}[section]
\newtheorem{proposition}{Proposition}[section]

\usepackage{algorithm}
\usepackage{algorithmic}
\usepackage{bbold}
\usepackage{multirow}
\usepackage{multicol}
\usepackage{makecell}

\title{Anomaly Detection with Adaptive and Aggressive Rejection for Contaminated Training Data}

\author{
  Jungi Lee, Jungkwon Kim, Chi Zhang, Kwangsun Yoo, Seok-Joo Byun\\
  ELROILAB Inc. \\
  Seoul, Republic of Korea\\
  \texttt{\{ganbbang12, jkkim, czhang, yks, sjbyun\}@elroilab.com} \\
}

\begin{document}
\maketitle

\begin{abstract}

Handling contaminated data poses a critical challenge in anomaly detection, as traditional models assume training on purely normal data. Conventional methods mitigate contamination by relying on fixed contamination ratios, but discrepancies between assumed and actual ratios can severely degrade performance, especially in noisy environments where normal and abnormal data distributions overlap. To address these limitations, we propose Adaptive and Aggressive Rejection (AAR), a novel method that dynamically excludes anomalies using a modified z-score and Gaussian mixture model-based thresholds. AAR effectively balances the trade-off between preserving normal data and excluding anomalies by integrating hard and soft rejection strategies. Extensive experiments on two image datasets and thirty tabular datasets demonstrate that AAR outperforms the state-of-the-art method by 0.041 AUROC. By providing a scalable and reliable solution, AAR enhances robustness against contaminated datasets, paving the way for broader real-world applications in domains such as security and healthcare.
\end{abstract}

% keywords can be removed

\section{Introduction}
\label{sec:intro}

Anomaly detection (AD) focuses on identifying previously unseen anomalies using only normal data. 
AD is widely applied in domains where anomalies are rare and difficult to obtain, such as cybersecurity~\cite{9152899,8538395}, fraud detection~\cite{9020760,HILAL2022116429}, and medical diagnosis~\cite{alloqmani2023anomaly,zhang2022unsupervised,10.1145/3464423}. 
However, in real-world applications, the assumption that datasets consist exclusively of normal data is often unrealistic, as abnormal data are frequently included during data collection.
Therefore, to improve practical applicability, it is crucial to make the models robust against contamination, ensuring consistent performance even in the presence of anomalous data.

In classification tasks, noisy labels are often identified based on discrepancies between predictions and assigned labels~\cite{englesson2021generalized, garg2023noisy, han2018co, jiang2020beyond, jiang2018mentornet, pleiss2020identifying, xu2019l_dmi, zhang2018generalized, lee2024fastsimifeat}. 
Other methods leverage semi-supervised learning techniques, incorporating cross-validation strategies and mix-up augmentation~\cite{cordeiro2023longremix, li2020dividemix}.
However, since AD typically involves training on a single normal class, it is inherently difficult to employ cross-entropy loss, which in turn limits the applicability of probability-based classification tasks~\cite{arazo2019unsupervised}. 
In prior studies involving contaminated datasets, a fixed contamination ratio -- typically set at 10\% -- was assumed.
Instances with high anomaly scores were excluded according to this preset ratio~\cite{yu2021normality, qiu2022latent}. 
However, this rigid rejection strategy introduces a trade-off between ``stability,'' which refers to retaining normal data in clean datasets, and ``robustness,'' which entails effectively excluding anomalies in contaminated datasets while minimizing the exclusion of normal data. 
A high rejection ratio enhances robustness by removing more anomalies but sacrifices stability as it may also eliminate a substantial portion of normal data.
Conversely, a low rejection ratio favors stability but compromises robustness by failing to filter out enough anomalous instances.

To address the issue, statistical outlier detectors can be employed to dynamically estimate the contamination ratio based on the distribution of anomaly scores~\cite{perini2023estimating, app112411854,bardet2017new}, rather than relying on a fixed ratio. 
These methods offer greater flexibility and can lead to more reliable and robust performance. 
However, their effectiveness diminishes when there is significant overlap between the distributions of normal and abnormal data, thus making the two difficult to distinguish between the two.
In such scenarios, relying on the assumption that anomaly scores follow a Gaussian distribution can further hinder accurate estimation of the contamination ratio. 
This highlights the need for more robust and assumption-free approaches to address these challenges. 

To overcome the limitations of traditional estimation methods, we propose Adaptive and Aggressive Rejection (AAR).
Instead of relying on a fixed contamination ratio, AAR dynamically identifies and excludes anomalies by using a modified z-score outlier detector~\cite{rousseeuw1993alternatives} applied to the anomaly score distribution within mini-batches during training.
To further enhance robustness in the presence of overlapping distributions, AAR incorporates an aggressive rejection strategy based on a Gaussian Mixture Model (GMM), enabling more effective removal of potential anomalies from normal data.
Rather than indiscriminately removing all suspected anomalies, -- which risks discarding informative normal samples -- we apply  a weighted loss function to the potential anomalies identified by our strategy. 
This design ensures that the model remains resilient to contamination while retaining valuable normal data for learning. 

This paper makes three key contributions: (1) We propose a dynamic contamination estimation method that eliminates the need for a fixed contamination ratio by leveraging a mini-batch-based statistical outlier detector.
(2) We theoretically and empirically demonstrate the importance of aggressive rejection in improving anomaly detection performance, especially under high distributional overlap.
(3) Building on these insights, we propose Adaptive and Aggressive Rejection (AAR), a novel approach that balances anomaly exclusion and normal data preservation through a weighted loss, resulting in a more stable and robust solution for anomaly detection.

\section{Related Work}\label{sec:related work}
\subsection{Anomaly Detection} 
Traditional anomaly detection methods, including one-class support vector machines (OC-SVM)~\cite{scholkopf1999support} and kernel density estimation~\cite{parzen1962estimation}, often struggle in high-dimensional spaces due to the curse of dimensionality. 
To address this, deep learning-based approaches have emerged, demonstrating significant advances in performance.
Early methods focused on reconstruction error-based models, including autoencoders (AE), variational autoencoders (VAE)~\cite{kingma2013auto}, and adversarial autoencoders (AAE)~\cite{makhzani2015adversarial}, which learn to reconstruct normal patterns while producing high reconstruction errors for anomalous inputs. 
The memory-augmented autoencoder (MemAE)~\cite{gong2019memorizing} introduced a memory module that maps the latent vector of the input to the most relevant latent vector stored in memory, enhancing anomaly detection performance. 
In a different line of work, deep support vector data description (DSVDD)~\cite{ruff2018deep} learns compact latent representations by constraining them within a hypersphere centered around a reference point.
Further improvements have been achieved via contrastive learning~\cite{reiss2023mean, tack2020csi}, outlier exposure~\cite{hendrycks2018deep}, and the inclusion of limited ground truth labels~\cite{ruff2019deep}. 
However, despite their success, most of these methods assume clean training data and are highly vulnerable to data contamination, which can distort learned representations and compromise detection performance.

\subsection{Anomaly Detection with Contaminated data}
Previous approaches have focused on designing robust models or loss functions for data refinement. 
For instance, the robust variational autoencoder with attention-based feature adaptation~\cite{gao2020rvae} builds upon the deep autoencoding Gaussian mixture model~\cite{zong2018deep}. 
They improve robustness by replacing the traditional autoencoder with a VAE and employing attention-based feature adaptation to balance the importance of the latent vector and reconstruction error. 
On the other hand, data refinement strategies aim to exclude samples identified as anomalies~\cite{gornitz2014learning, xia2015learning, yoon2021self}. 
For example, iterative training set refinement~\cite{beggel2019robust} integrates OC-SVM~\cite{scholkopf1999support} with the latent space of an AAE to iteratively refine the training set.
Similarly, the normality-calibrated autoencoder~\cite{yu2021normality} identifies high-confidence normal samples within a low-entropy space and leverages them to enhance anomaly prediction.
While effective, these methods typically require modifications to the underlying models, limiting their adaptability and generalizability across different architectures.

Robust loss functions that enhance model robustness without requiring architectural modifications offer an alternative approach to anomaly detection. 
The pseudo-Huber loss (Huber)~\cite{liznerski2021explainable}, a classical loss function widely applied across various domains, replaced the standard L2 loss in anomaly detection tasks. 
However, while they enhance robustness by lowering the gradient for anomalies, they exhibit suboptimal performance due to the incomplete exclusion of abnormal instances. 
Latent outlier exposure (LOE)~\cite{qiu2022latent} introduces separate loss functions for normal and abnormal instances, balancing them by targeting the top 10\% of samples with the lowest confidence. 
Similarly, iterative anomaly detection (IAD)~\cite{kim2023iterative} uses an adaptive weighting mechanism that iteratively refines the learning process based on the inferred normality of samples. 
LOE heavily relies on a predefined contamination ratio and the design of its abnormal loss function. 
The re-weighting strategy of IAD, which treats all samples as potentially anomalous, limits the efficacy.

\subsection{Contamination Ratio Estimation} 
Outlier detection techniques can be leveraged for the dynamic threshold, enabling the exclusion of potential anomalies within mini-batches without relying on a fixed contamination ratio. For instance, the inter-quartile range (IQR)~\cite{bardet2017new} and quasi-monte carlo discrepancy (QMCD)~\cite{iouchtchenko2019deterministic} estimate the contamination ratio based on quantiles. The modified thompson tau test~\cite{rengasamy2021towards} estimates the contamination ratio through statistical inference. The $\gamma$ Gaussian mixture model ($\gamma$GMM)~\cite{perini2023estimating} calculates the posterior distribution of contamination using various anomaly scores, including local outlier factor (LOF)~\cite{breunig2000lof}, $k$-nearest neighbors ($k$NN)~\cite{angiulli2002fast}, and isolation forest (IForest)~\cite{liu2012isolation}. Although these estimations perform effectively on tabular datasets, they frequently perform inferior on image datasets. Furthermore, even with precise contamination estimation, the inherent similarity between normal and abnormal samples can make it difficult to eliminate all anomalous data.

\section{Proposed Method}
\label{sec:method}
In this section, we introduce the proposed AAR, a novel framework that integrates an outlier detector with GMM.
We begin by establishing the necessity of aggressive sample rejection from both the theoretical and empirical point of view,  demonstrating its impact on the separation between normal and abnormal distributions. 
Building upon this insight, we formalize AAR by presenting key implementation details which enables AAR to balance robustness and stability.

\subsection{Preliminaries}
In traditional anomaly detection, it is typically assumed that the training dataset is clean, consisting solely of normal instances. 
Let this clean dataset be denoted by $D_n:=\{(x_i,y_i)\}^{N_n}_{i=1}$, where $y_i = 0$ indicates a normal sample and $N_n$ is the number of such instances.
However, in real-world scenarios, datasets are often contaminated by abnormal instances. 
Let the set of abnormal samples be denoted by $D_a:=\{(x_i,y_i)\}^{N_a}_{i=1}$, where $y_i=1$ indicates an anomaly and $N_a$ is the number of such samples.
As a result, the observed dataset is the union $\mathcal{D}= D_n \cup D_a$, consisting of both normal and abnormal instances. 
The objective is to effectively identify and exclude the anomalies in $D_a$, so as to learn a model using only the normal instances in $D_n$.

To measure the robustness of a rejection threshold based on the quantile of anomaly scores, we define the following function with respect to the loss function:
\begin{definition} [Robustness of a rejection threshold]\label{def:robustness} 
Let $\mathbf{z}:=(x_i, y_i) \in \mathcal{D}$ be a sample and $s(\mathbf{z})$ be an anomaly score such as the reconstruction error $\|x_i -f(x_i)\|^2_2$, and let $s_q(\mathbf{z})$ be the $q$-quantile of $s(\mathbf{z})$ over $\mathcal{D}$.
For a class label $\ell \in \{0, 1 \}$, we define the subset $I :=\{ \mathbf{z} : s_q(x_i) \in [s_{q_1}, s_{q_2}], y_i = \ell\}$.
Then $L(q_1, q_2,\ell) = \sum_{ \mathbf{z}\in \mathcal{D}} \mathbb{1}_{I}(\mathbf{z}) s(\mathbf{z})$ denotes the cumulative sum of scores $s(\mathbf{z})$ in the quantile range $[q_1, q_2]$.
The robustness of a rejection threshold at $q$-quantile is then defined as 
\begin{equation}\label{eq:robustness}
    R(q)  =\frac{L(0,q,0)}{L(0,1,0)}-\frac{L(0,q,1)}{L(0,1,1)}
\end{equation}
which quantifies the difference between relative anomaly scores of normal ($\ell=0$) and abnormal ($\ell=1$) instances below the $q$-quantile.
\end{definition}

% \begin{figure}[t]
%   \centering
%   \subfigure[Effect of aggressive rejection.]
%   {
%     \includegraphics[width=0.31\textwidth]{Figure1a.pdf}
%     \label{fig:1a}
%   }
%   \hfill
%   \subfigure[Distribution of the MNIST class 6.]
%   {
%     \includegraphics[width=0.31\textwidth]{Figure1b.pdf}
%     \label{fig:1b}
%   }
%   \hfill
%   \subfigure[Robustness across quantiles.]
%   {
%     \includegraphics[width=0.31\textwidth]{Figure1c.pdf}
%     \label{fig:1c}
%   }
%   \caption{
%   Motivation for aggressive rejection. (a) demonstrates that a 5–10\% higher rejection ratio $\gamma$ yields better performance than using the true contamination ratio ($\gamma=\gamma_0$). (b) illustrates the distribution of anomaly scores of normal and abnormal samples. (c) presents the robustness across quantiles. Notably, while a 0.9-quantile rejection achieves a robustness of 0.63 and an AUROC of 0.928, a more aggressive 0.8-quantile rejection outperforms it with a higher robustness of 0.69 and an AUROC of 0.936.
%   }
%   \label{fig:1}
% \end{figure}

\begin{figure}
    \centering
    \begin{subfigure}{0.33\linewidth}
        \includegraphics[width=\textwidth]{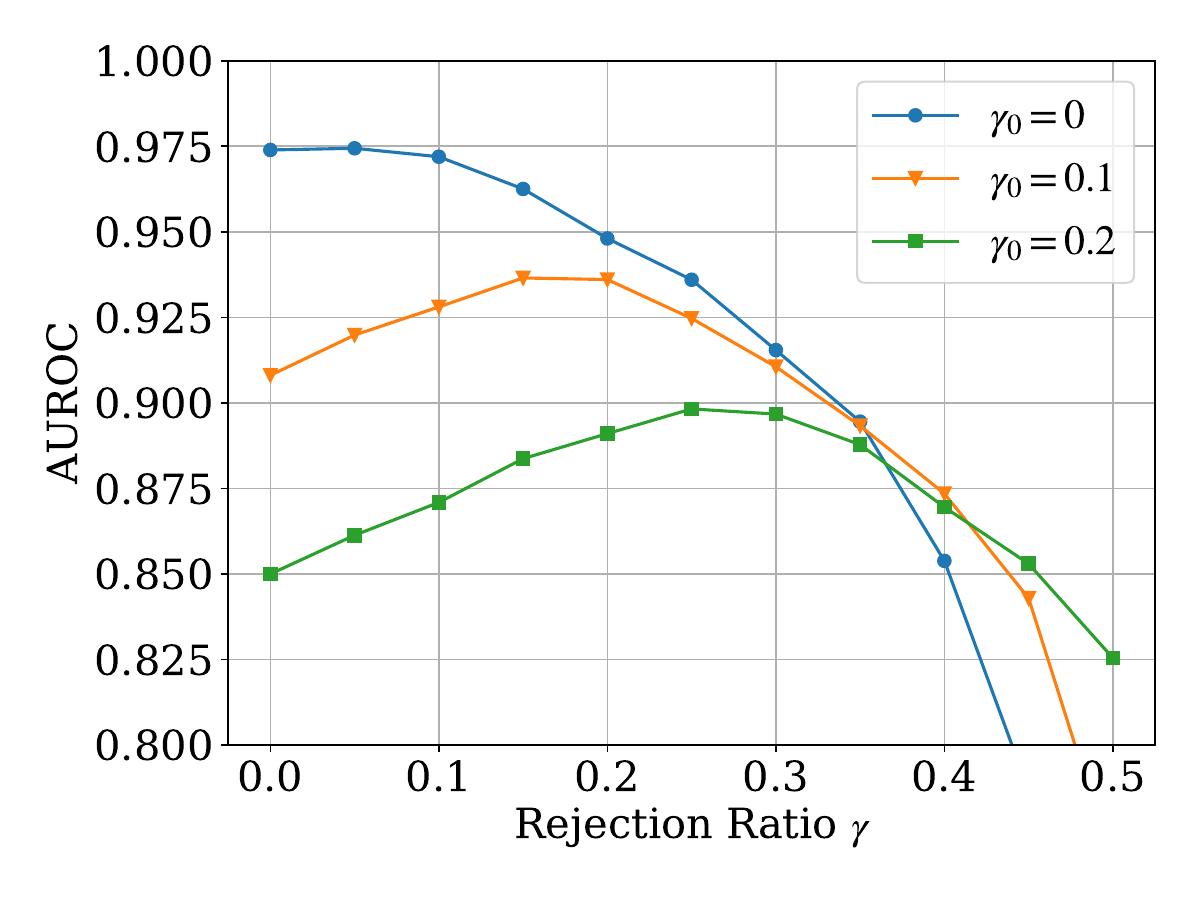}
        \caption{Effect of aggressive rejection.}
        \label{fig:1a}
    \end{subfigure}
        \begin{subfigure}{0.33\linewidth}
        \includegraphics[width=\textwidth]{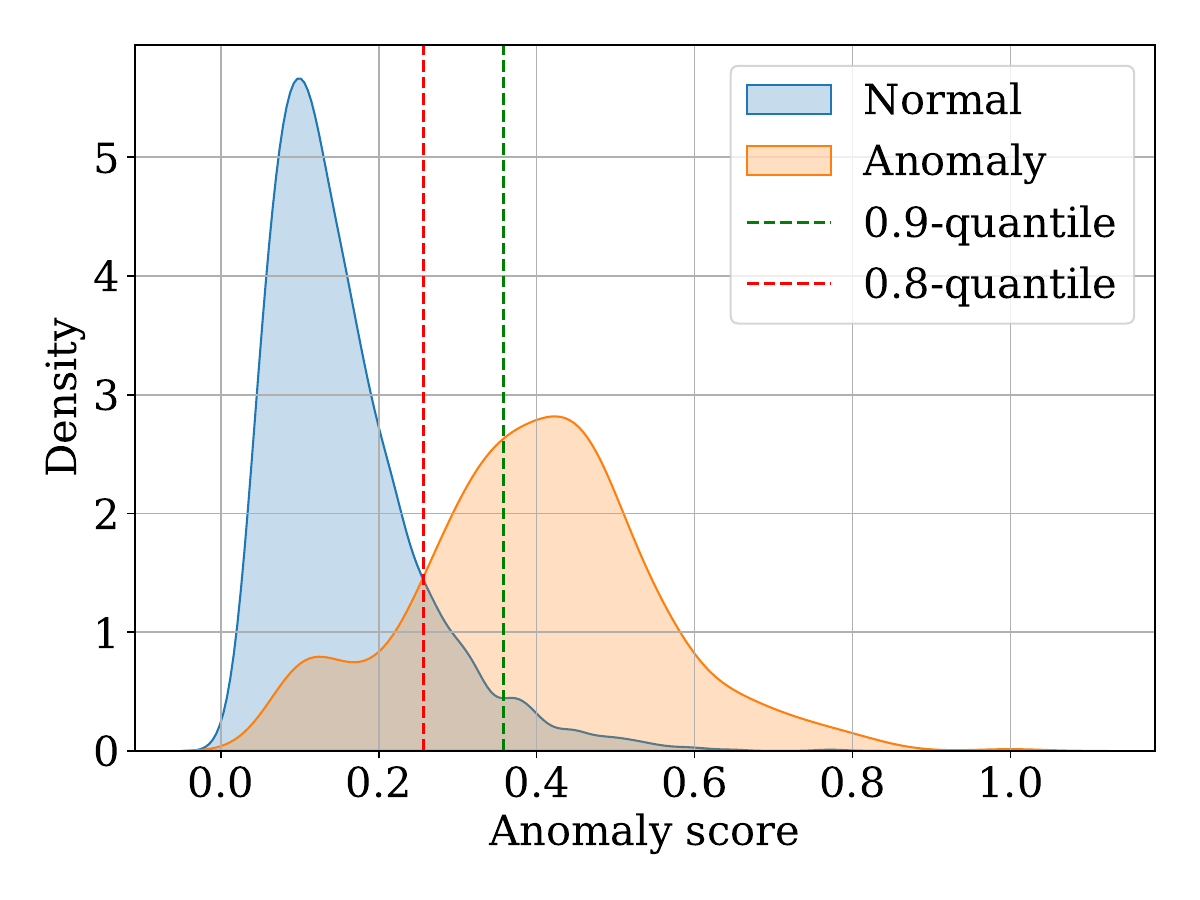}
        \caption{Distribution of the MNIST class 6.}
        \label{fig:1b}
    \end{subfigure}
        \begin{subfigure}{0.33\linewidth}
        \includegraphics[width=\textwidth]{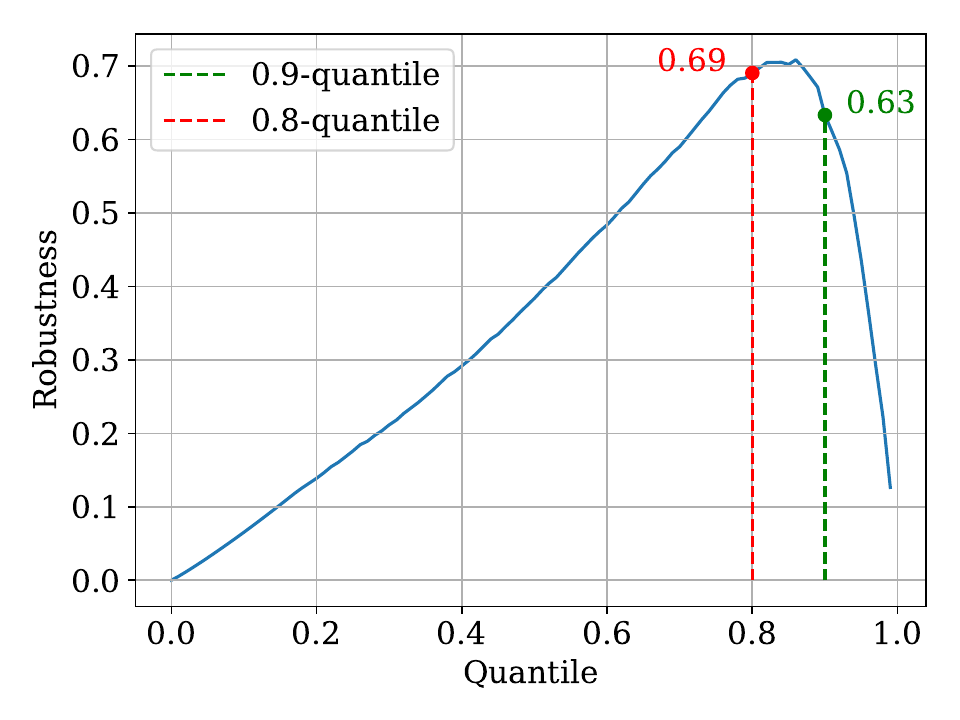}
        \caption{Robustness across quantiles.}
        \label{fig:1c}
    \end{subfigure}
    \caption{
    Motivation for aggressive rejection. (a) demonstrates that a 5–10\% higher rejection ratio $\gamma$ yields better performance than using the true contamination ratio ($\gamma=\gamma_0$). (b) illustrates the distribution of anomaly scores of normal and abnormal samples. (c) presents the robustness across quantiles. Notably, while a 0.9-quantile rejection achieves a robustness of 0.63 and an AUROC of 0.928, a more aggressive 0.8-quantile rejection outperforms it with a higher robustness of 0.69 and an AUROC of 0.936.
    }
    \label{fig:1}
\end{figure}

\subsection{Motivation}
The contamination ratio is typically set to 10\% to balance robustness and stability. 
However, we observe that using a contamination ratio 5–10\% higher than the true contamination ratio improves robustness due to the overlap between normal and abnormal distributions. 
Figure~\ref{fig:1a} illustrates performance on a contaminated MNIST dataset~\cite{lecun2010mnist} (class 6) with varying rejection ratios.
Then we can find a more effective threshold for maximizing $R(q)$ in \eqref{eq:robustness} using the following proposition:
\begin{proposition} \label{theorem:overestimation}
Let $s_n(x)$ and $s_a(x)$ be the probability density functions on anomaly scores of normal and anomaly samples, respectively.
Suppose the overall distribution of anomaly scores is a mixture of the two distributions with ratio $\alpha \in (0, 1)$ such that $S(x) = \alpha s_n(x)+(1-\alpha)s_a(x)$.
Then $R(q)$ in \eqref{eq:robustness} has its maximum at $q^{\ast}$-quantile satisfying 
\begin{equation}\label{eq1}
    \frac{s_n(\tau_{q^{\ast}})}{s_a(\tau_{q^{\ast}})} = \frac{\mathbb{E}[s_n]}{\mathbb{E}[s_a]}, \quad \textnormal{for} \quad 
     q^{\ast} = \int_0^{\tau_{q^{\ast}}} S(x) dx.
\end{equation}
\end{proposition}

\begin{proof}
Using the definition of $L(q_1, q_2, \ell)$, we see
\begin{equation}
    R(q) = \frac{N_n\int_0^{\tau_q} x s_n(x) dx}{N_n \mathbb{E}[s_n]} - \frac{N_a \int_0^{\tau_q} x s_a(x) dx}{N_a \mathbb{E}[s_a]}.
\end{equation}
Replacing $q^{\ast}$ with $q$ in the second equality in \eqref{eq1} and taking derivative with regard to $q$, we also have
\begin{equation}\label{eq2}
        \frac{d\tau_q}{dq} = \frac1{S(\tau_q)}.        
\end{equation}
So, the derivative of $R(q)$ is given by 
\begin{equation}
    \begin{aligned}
    \frac{dR(q)}{dq} & = \tau_q \Big(\frac{s_n(\tau_q)}{\mathbb{E}[s_n]} - \frac{s_a(\tau_q)}{\mathbb{E}[s_a]}\Big)\frac{d\tau_q}{dq},  \\
    & = \frac{\tau_q}{S(\tau_q)}\Big(\frac{s_n(\tau_q)}{\mathbb{E}[s_n]} - \frac{s_a(\tau_q)}{\mathbb{E}[s_a]} \Big).
    \end{aligned}
\end{equation}
Hence, $R(q)$ has its critical points at $q=q^{\ast}$ satisfying the conditions in \eqref{eq1}.
Now, it suffices to show $d^2R(q)/dq^2 < 0$ at $q^{\ast}$.
For simplicity, let $D(\tau_q) = s_n(\tau_q) / \mathbb{E}[s_n] - s_a(\tau_q) / \mathbb{E}[s_a]$.
Then the second derivative of $R(q)$ is expressed as 
\begin{equation}
    \frac{d^2R(q)}{dq^2} = \frac{D(\tau_q)S(\tau_q) + \tau_q D'(\tau_q)S(\tau_q) - \tau_qD(\tau_q)S'(\tau_q)}{\{S(\tau_q)\}^3},
\end{equation}
which is negative at $q=q^{\ast}$ if and only if $D'(q^{\ast}) <0$ since $D(\tau_{q^{\ast}}) = 0$, $\tau_{q^{\ast}} >0$ and $S(\tau_{q^{\ast}})>0$.
In practice, the distribution of anomaly scores for normal samples lies on the left of that for abnormal samples with $s_{n}'(\tau_{q^{\ast}}) < 0 < s_{a}'(\tau_{q^{\ast}})$. 
This implies $D'(\tau_{q^{\ast}}) <0$ which completes the proof.     
\end{proof}

We note here that the Proposition~\ref{theorem:overestimation} suggests the aggressive rejection can be more effective when the distributions of normal and abnormal samples are less distinguishable.
While the conclusion is derived under certain assumptions that may not fully capture all real-world scenarios -- such as variations in data characteristics or training dynamics -- it indicates that, as the two distributions become more similar, the ratio $\mathbb{E}[s_n]/\mathbb{E}[s_a]$ tends to increase.
This in turn reduces the required threshold $\tau_{q^{\ast}}$ for which the first condition in \eqref{eq1} holds, making aggressive rejection more likely to reject ambiguous samples.
Such behavior may be particularly beneficial in challenging scenarios where distinguishing between normal and abnormal data is inherently difficult.

From these observations, we conclude that the thresholding scheme based on the loss function prioritizes the exclusion of abnormal instances while minimizing the impact on normal instances by adopting a higher rejection ratio. 
Notably, setting a higher rejection ratio can enhance robustness when the reduction in loss ratio for abnormal samples exceeds that for normal instances. 
As shown in Figure~\ref{fig:1b} and ~\ref{fig:1c}, increasing the rejection ratio by  10\% results in improved robustness. 
Specifically, this adjustment results in a 0.06 increase in robustness and a 0.008 improvement in AUROC, outperforming the baseline with a fixed 10\% rejection ratio.

Our theoretical and empirical analyses jointly underscore the importance of adopting a relatively high rejection ratio to achieve effective anomaly detection. 
However, excessively high rejection ratios risk excluding normal data, potentially deteriorating the robustness. 
This trade-off highlights the necessity for an adaptive approach rather than a fixed rejection ratio -- one enables dynamic adjustments to varying levels of contamination in the data. 
To address this challenge, we propose the AAR method, which integrates an outlier detector and a GMM.

\subsection{Adaptive and Aggressive Rejection} \label{sec:method_AAR}
To effectively suppress the influence of anomalous samples throughout training, we adopt an adaptive rejection strategy that evolves with the model's learning progress.
In the early phase, when the model is still unstable, we apply a hard rejection method to ensure that training focuses on the most trustworthy normal samples.
This stabilizes optimization and prevents the model from being misled by potential anomalies.
As training advances and the anomaly scoring becomes more reliable, we progressively combine hard and soft rejection schemes.
This transition allows the model to handle more ambiguous cases with refined precision, enabling both robust early learning and flexible late-stage adaptation.

\paragraph{Hard rejection with normality threshold}
To implement hard rejection in the early phase, we leverage a modified z-score (MZ) based outlier detector~\cite{rousseeuw1993alternatives} that adaptively filters out anomalies within each mini-batch.
The MZ method replaces the mean and standard deviation with the median and normalized median absolute deviation (MAD), making it more robust against extreme outliers.
For anomaly scores $s_i$ of a sample $(x_i, y_i) \in \mathcal{D}$, 
the modified z-score $m_i$ is defined as follows:
\begin{equation}\label{eq:mz-score}     
\begin{gathered}
    \hat{s} = \textnormal{median}_{i} (s_i), \\
    \textnormal{MAD} = \textnormal{median}_{i} |s_i-\hat{s}|, \\
    m_i= 0.6745(s_i-\hat{s})/\textnormal{MAD}.
\end{gathered}
\end{equation}
Using this, we derive the normality threshold $\tau_N$ as
\begin{equation}
    \tau_N= 3.5 \times \textnormal{MAD}/0.6745 + \hat{s},
    \label{eq:Tn}
\end{equation}
where 3.5 is a widely used constant for outlier detection~\cite{iglewicz1993volume}. 
Instances with scores exceeding $\tau_N$ are considered outliers and excluded from training.
This adaptive criterion allows the rejection mechanism to flexibly adjust to the current distribution of scores, making it effective across varying contamination levels without requiring manual tuning.

\begin{figure}[t]
  \centering
  \includegraphics[width=\textwidth]{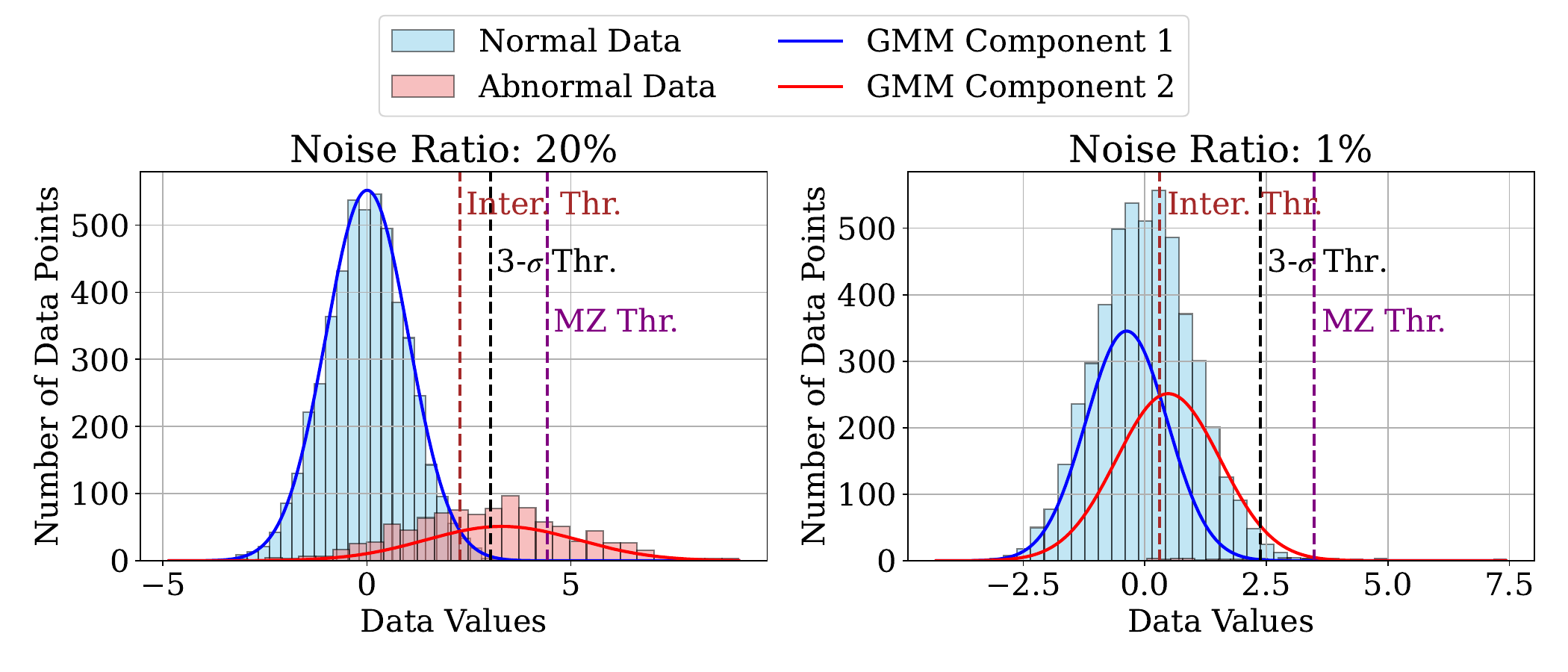}
   \caption{Thresholds applied to two Gaussian models with contamination ratios of 20\% (left) and 1\% (right) demonstrate significant variations in performance. 
   For datasets with a high contamination ratio of 20\%, the intersection threshold which is defined as the point where normal and abnormal distributions overlap effectively excludes anomalies, thereby optimizing robustness. 
   Conversely, the MZ threshold proves inadequate, failing to sufficiently exclude all anomalies. 
   However, on datasets with a low contamination ratio of 1\%, the intersection threshold excludes numerous normal data points, adversely affecting stability. 
   In such cases, the $z$-$\sigma$ threshold (e.g., $z=3$) offers a balanced solution, achieving superior robustness compared to the MZ threshold while providing greater stability than the intersection threshold. 
   This balance makes the $z$-$\sigma$ threshold a more reliable and versatile choice for handling datasets with varying levels of contamination.}
   \label{fig:2}
\end{figure}

\paragraph{Soft Rejection with employing GMM} 
As the model becomes more stable in later stages of training, we incorporate a soft rejection mechanism that enables nuanced decisions beyond hard thresholding, which shows the binary nature.
While the MZ threshold provides a robust basis for hard rejection, it tends to be overly conservative -- especially under high contamination -- due to the long-tailed nature of the anomaly score distribution.
This can result in insufficient exclusion of low-confidence anomalies.
Figure~\ref{fig:2} illustrates the MZ threshold (purple line) under contamination ratios of 20\% and 1\%, with samples drawn from a Gaussian distribution. 
Ideally, the threshold should be set at the intersection point between the normal and abnormal distributions, as defined by Proposition~\ref{theorem:overestimation}. 
However, the MZ threshold tends to lie above this optimal boundary, failing to effectively reject ambiguous or low-confidence anomalies.
To overcome this, we propose a soft rejection method based on GMM. 
We assume that the distribution of anomaly scores can by approximated by a mixture of two Gaussian components -- one representing normal samples and the other capturing anomalies.
By fitting a GMM to the current anomaly score distribution, we can estimate the parameters of these components and analytically compute their intersection point, denoted by $\tau_I$.
This intersection approximates the optimal threshold that separates normal and abnormal regions in the score space.
A GMM is a probability density function that can be expressed as a weighted sum of $k$ Gaussian distributions as follows:
\begin{equation}
        p(x) = \sum_{k=1}^K \pi_k \cdot \mathcal{N}(x \mid \mu_k, \sigma_k^2), \quad 0 \leq \pi_k \leq 1, \quad \sum_{k=1}^K \pi_k = 1,
\end{equation}
where $x$ is data, $K$ is the number of Gaussian components, $\pi_k$ is a mixing coefficient for $k$-th component, and $\mathcal{N}(x \mid \mu_k, \sigma_k^2)$ is a Gaussian distribution with mean $\mu_k$ and variance $\sigma_k^2$. 
We set $k$ to 2, as we only consider normal and abnormal distributions.
The intersection, where the probability density functions of the two Gaussians are equal ($\mathcal{N}(\tau_{I} \mid \mu_1, \sigma_1^2)=\mathcal{N}(\tau_{I} \mid \mu_2, \sigma_2^2), \, \mu_1 < \tau_I < \mu_2$), is given by one of 
\begin{equation}
    \tau_{I}=\frac{-b \pm \sqrt{b^2 - ac}}{a},
    \label{eq:Ti}
\end{equation}
where $a = \frac{1}{\sigma_1^2} - \frac{1}{\sigma_2^2}$, $b = \frac{\mu_2}{\sigma_2^2} - \frac{\mu_1}{\sigma_1^2}$, and $c = \frac{\mu_1^2}{\sigma_1^2} - \frac{\mu_2^2}{\sigma_2^2} - 2\ln\frac{\sigma_2}{\sigma_1}$. 
However, directly relying on $\tau_I$ can be problematic in practice. 
When the contamination ratio is negligible, as illustrated in the right panel in Figure~\ref{fig:2}, the intersection threshold is set too aggressively -- discarding a significant portion of normal data.
On the other hand, in cases where the clean dataset exhibits a long-tailed distribution, the standard intersection threshold results in excessive normal data loss. 
In either case, relying solely on the threshold $\tau_I$ can compromise the model's stability.
To mitigate this, we complement $\tau_I$ with a secondary threshold $\tau_\sigma$, which can be applied when the distribution of normal data is known, defined as
\begin{equation}
    \tau_{\sigma} = z \cdot \sigma_n+\mu_n.
    \label{eq:Tsigma}  
\end{equation}
Here $\mu_n$ and $\sigma_n$ are the mean and standard deviation of the normal distribution, respectively. 
Also, $z$ is a hyperparameter that indicates how far the threshold deviates from the mean of the distribution, in units of standard deviation.
As illustrated in Figure~\ref{fig:2}, while the threshold approaches the intersection threshold on a high contamination ratio, it aligns closely with the MZ threshold on a low ratio.
Also, the $z$-$\sigma$-based threshold strikes a balance between noise rejection and data preservation, making it particularly effective in cases where long-tailed distributions are present.

In our method, we define the final soft rejection threshold as $\tau = \max(\tau_\sigma, \tau_I)$.
Samples with scores between $\tau_s$ and the MZ threshold $\tau_N$ are softly rejected by assigning them a smaller weight $t_s$ during training.
This weighted penalty allows the model to remain cautious about uncertain cases without completely discarding them.
By integrating both GMM-based and distribution-aware thresholds, our approach balances aggressive anomaly exclusion and the preservation of informative normal samples.

\begin{algorithm}
   \caption{Adaptive and Aggressive Rejection for Robust anomaly detection}
   \label{alg:main}
\begin{algorithmic}
   \STATE  {\bfseries Input:} Sample $X$, model $f$, hyperparameters, $E$, $z$, $t_s$
   \FOR{\textbf{each} epoch}
       \FOR{\textbf{each} Mini-batch $\bf{x} \subseteq \bf{X}$}
           \STATE $\textbf{s}=||\textbf{x}-f(\textbf{x})||^2_2$ \hfill // Calculate anomaly scores
           \STATE Set Normality Threshold $\tau_N$ \hfill // Equation~\eqref{eq:Tn}
           \STATE Fit Gaussian Mixture Model with $\bf{s}$
           \STATE Set Intersection Threshold $\tau_{I}$ \hfill //{Equation~\eqref{eq:Ti}}
           \STATE Set $z$-$\sigma$ Threshold $\tau_{\sigma}$  \hfill   // Equation~\eqref{eq:Tsigma}
           \STATE $\tau = \text{max}(\tau_{\sigma}, \tau_{I})$

           \IF{$\text{epoch} \le E$}
               \STATE $w_i = 
                    \begin{cases}
                    0, & \text{if  } s_i > \tau_{N},\\
                    1, & \text{otherwise.}
                    \end{cases}
                    $
           \ELSE
               \STATE $w_i = 
                    \begin{cases}
                    0, & \text{if  } s_i > \tau_{N},\\
                    t_s, & \text{if  } \tau < s_i \leq \tau_{N},\\
                    1, & \text{otherwise.}
                    \end{cases}
                    $
           \ENDIF

            \STATE $L = \frac{1}{N}\sum^N_{i=1}w_i \cdot ||x_i-f(x_i)||^2_2$ 
            \STATE Update model parameters with $L$
       \ENDFOR
   \ENDFOR
\end{algorithmic}
\end{algorithm}

\subsection{Algorithm}
In the early stages of training, we train a model with a warm-up training phase that trains the model on instances with anomaly scores lower than the MZ threshold $\tau_N$ for $E$ epochs as the anomaly scores are randomly scattered. Next, we fit a GMM to the anomaly scores and set two thresholds, $\tau_{I}$ and $\tau_\sigma$. When the normal and abnormal distributions are completely separated, $\tau_{I}$ may exceed $\tau_{\sigma}$. Therefore, the lower bound of $\tau_{\sigma}$ is set to $\tau_{I}$. We assign a weight $w_i=0$ for hard rejection and $w_i=t_s$ for soft rejection. Finally, the model is trained using a weighted loss function.

%-------------------------------------------------------------------------
%Evaluation 

\section{Evaluation}
\label{evaluation}
This section compares existing sample selection methods and robust loss functions with our methods. Two fundamental image datasets-MNIST~\cite{lecun2010mnist}, FashionMNIST (F-MNIST)~\cite{DBLP:journals/corr/abs-1708-07747}-and thirty tabular datasets~\cite{Rayana:2016,Dua:2019} are used to evaluate the methods. MNIST and F-MNIST consist of 10 classes and 28$\times$28 gray scale images. Thirty tabular datasets include multi-dimensional point datasets such as healthcare and cybersecurity, as in LOE~\cite{qiu2022latent}.

\subsection{Datasets and setups}
A one-vs-rest setup is used for image datasets, where one class is considered normal and the other classes are treated as abnormal~\cite{gong2019memorizing,ruff2018deep,ruff2019deep,zong2018deep}. We add $\gamma/(1-\gamma)*N$ abnormal data, where $\gamma$ is the contamination ratio and $N$ is the number of normal data. The area under receiver operating characteristic (AUROC) is used as the evaluation metric. In the experiments, each class is set as normal, and the average AUROC is measured using three different random seeds. For tabular datasets, half of the normal samples are used for training. Due to the absence of anomalies, artificial anomalies for contaminated training data are generated by adding zero-mean Gaussian noise to test anomalies, as in Shenkar and Wolf ~\cite{shenkar2021anomaly} and Qiu~\textit{et al.}~\cite{qiu2022latent}. The standard deviations for the contamination are derived from test anomalies. We measure the average AUROC with ten different seeds on each dataset. Additionally, we leverage a Gaussian mixture model with two components and set $E$, $z$, and $t_s$ as 15, 2.5, and 0.1, respectively.

\subsection{Comparison Methods}

We use three distinct baseline models to ensure broad applicability. AE~\cite{hinton2006fast,bergmann2018improving} is a conventional reconstruction-based model, while MemAE~\cite{gong2019memorizing} is an AE-based model that incorporates a memory module and an additional loss function. DSVDD~\cite{ruff2018deep} is a one-class classification model with a training process distinct from AE.

We employ various sample selection strategies and robust loss functions for the comparison methods. The sample selection methods include Inter-Quartile Range (IQR)~\cite{bardet2017new}, modified z-score (MZ)~\cite{rousseeuw1993alternatives,yaro2024outlier}, Quasi-Monte Carlo Discrepancy outlier detection (QMCD)~\cite{iouchtchenko2019deterministic}, $\gamma$GMM~\cite{perini2023estimating}, and fixed 10/20\% contamination ratio approach. IQR uses the threshold 
Q3+1.5(Q3 - Q1), where Q1 and Q3 are the first and third quartiles, respectively. QMCD employs the quantile derived from one minus the quasi-monte carlo discrepancy. MZ is a variation of the z-score designed to be robust to outliers, where z-scores exceeding 3.5 are typically considered outliers. $\gamma$GMM estimates the posterior distribution of the contamination factor. Except for $\gamma$GMM, all other methods are applied to mini-batches to remove detected outliers during training. In contrast, $\gamma$GMM utilizes scores from multiple outlier detectors to measure the contamination ratio, which is determined before training. Although $\gamma$GMM adapts dynamically based on the dataset characteristics, it relies on a predetermined contamination ratio.

We employ Huber~\cite{huber1992robust, liznerski2021explainable}, LOE~\cite{qiu2022latent}, and IAD~\cite{kim2023iterative} as existing robust loss functions. Huber mitigates the gradient impact of high anomaly scores in MSE. LOE balances normal and abnormal samples by applying the method specifically to the top 10\% of samples with high anomaly scores. IAD typically employs fixed weights across training iterations, but for evaluation as a loss function, we modify it by setting weights dynamically within each batch.

\subsection{Validation of Aggressive Rejection}

\begin{figure}[t]
  \centering
    \includegraphics[width=0.99\textwidth]{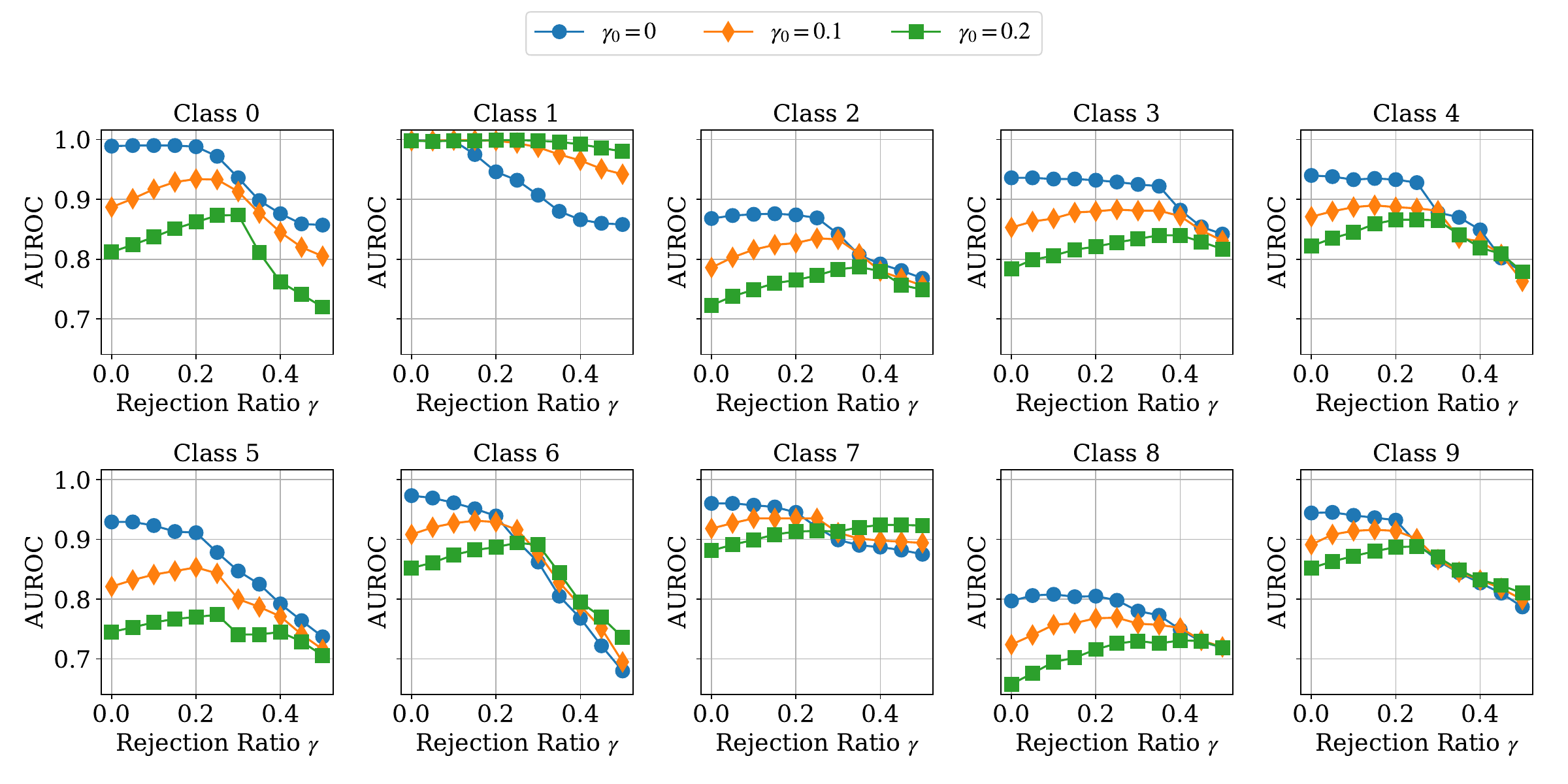}   
  \caption{Validation of aggressive rejection with various rejection ratio on MNIST dataset.
  }
  \label{fig:5}
\end{figure}

To validate the effect of aggressive rejection, we conduct experiments on the MNIST dataset using various rejection ratios, as shown in Figure~\ref{fig:5}. By systematically increasing the rejection ratio beyond the true contamination ratio, we evaluate its impact on robustness and anomaly detection performance. Our experiments demonstrate that overestimating the contamination ratio by 5-10\% enhances robustness and AUROC. Specifically, as the rejection ratio increases, the proportion of abnormal instances in the training set decreases, resulting in a more stable and reliable model. However, an excessively high rejection ratio may exclude a substantial portion of normal data, potentially reducing overall performance. These results align with our theoretical findings, reinforcing the necessity of aggressive rejection in anomaly detection.

\subsection{Experiments on Image Datasets}
\paragraph{Sample Selection}
We compare various sample selection methods, including IQR, MZ, QMCD, $\gamma$GMM, and the fixed 10/20\% removal (denoted as 10/20\% Rej.), to evaluate the performance of anomaly detection in the presence of contamination. Table~\ref{tab:sampleselection} compares the results on two image datasets—MNIST and F-MNIST—under 0\% (clean) and 10\% contamination scenarios. AAR consistently outperforms all other methods across datasets by 0.004-0.03, demonstrating the best balance between normal and contaminated data. MZ, IQR, QMD, $\gamma$GMM, and 10\% fixed rejection show degraded performance on clean datasets due to the long-tailed distributions but improve robustness at 20\% contamination. The 20\% rejection achieves the highest robustness among comparison methods, but its excessive removal of normal data leads to a significant drop in performance. In contrast, AAR maintains a higher stability than the 20\% rejection, whereas achieves comparable or superior robustness across the existing methods. 

\begin{table}
% \small
\caption{AUROC of sample selection methods on image datasets. $\gamma_0=0$ and $\gamma_0=0.2$ mean a clean and contaminated dataset, respectively. Average (Avg.) indicates the average AUROC between $\gamma_0=0$ and $\gamma_0=0.2$, meaning the balance of stability and robustness.}
\label{tab:sampleselection}
\begin{center}

\begin{tabular}{@{\extracolsep\fill}ccccccccccccc}
\hline
Dataset   &Model& $\gamma_0$   & MSE & 10\% Rej. & 20\% Rej. & IQR & MZ & QMCD & $\gamma$GMM  & AAR\\
\hline
\multirow{9}{*}{MNIST}
& \multirow{3}{*}{AE}

&0 & \textbf{0.933} & 0.932 & 0.920 & \textbf{0.933} & \textbf{0.933} & \textbf{0.933}  & 0.929 &  0.929 \\ 
&&0.2 & 0.813 & 0.834 & 0.849 & 0.821 & 0.819 & 0.823  & 0.841 & \textbf{0.850} \\ 
&&Avg. & 0.873 & 0.883 & 0.885 & 0.877 & 0.876 & 0.878  & 0.885 & \textbf{0.890} \\ 
\cline{2-11}

& \multirow{3}{*}{MemAE}
&0 & \textbf{0.928} & 0.924 & 0.882 & 0.926 & 0.925 & \textbf{0.928} & 0.910  & 0.923 \\ 
&&0.2 & 0.780 & 0.809 & 0.819 & 0.802 & 0.789 & 0.799 &0.815  & \textbf{0.833} \\ 
&&Avg. & 0.854 & 0.867 & 0.851 & 0.864 & 0.857 & 0.863 & 0.863& \textbf{0.878} \\ 
\cline{2-11}

& \multirow{3}{*}{DSVDD}
&0 & \textbf{0.927} & 0.918 & 0.902 & 0.920 & 0.919 & 0.920 & 0.915 & 0.911 \\ 
&&0.2 & 0.797 & 0.819 & \textbf{0.839} & 0.817 & 0.821 & 0.816  & 0.822 & 0.838 \\ 
&&Avg. & 0.862 & 0.869 & 0.870 & 0.868 & 0.870 & 0.868  & 0.869 & \textbf{0.874} \\ \hline 

\multirow{9}{*}{F-MNIST}
& \multirow{3}{*}{AE}
&0 & 0.884 & 0.885 & 0.870 & 0.886 & 0.886 & 0.885  & 0.884 & \textbf{0.887} \\ 
&&0.2 & 0.788 & 0.808 & 0.828 & 0.804 & 0.806 & 0.799  & 0.812 & \textbf{0.835} \\ 
&&Avg. & 0.836 & 0.846 & 0.849 & 0.845 & 0.846 & 0.842  & 0.848& \textbf{0.861} \\  
\cline{2-11}

& \multirow{3}{*}{MemAE}
&0 & 0.888 & 0.885 & 0.855 & 0.892 & 0.892 & 0.890  & 0.883 & \textbf{0.893} \\ 
&&0.2 & 0.763 & 0.793 & \textbf{0.818} & 0.785 & 0.786 & 0.793  & 0.797 & 0.809 \\ 
&&Avg. & 0.826 & 0.839 & 0.837 & 0.838 & 0.839 & 0.841  & 0.840 & \textbf{0.851} \\ 
\cline{2-11}

& \multirow{3}{*}{DSVDD}
&0 & \textbf{0.924} & 0.922 & 0.910 & 0.918 & 0.921 & 0.920  & 0.922 & 0.919 \\ 
&&0.2 & 0.817 & 0.839 & \textbf{0.877} & 0.842 & 0.855 & 0.837  & 0.854 & 0.876 \\ 
&&Avg. & 0.871 & 0.881 & 0.894 & 0.880 & 0.888 & 0.878  & 0.888 & \textbf{0.897} \\ 

\hline

\end{tabular}
\end{center}
\end{table}

\paragraph{Robust Loss Function}
We evaluate our methods alongside existing robust loss functions, including Huber, LOE, and IAD. Experiments are conducted on both clean and 10\% contaminated datasets, as LOE has been reported to achieve the best performance when the true and assumed contamination ratios align ($\gamma_0=\gamma=0.1$). Table~\ref{tab:robustloss} presents the results on two image datasets. Huber loss adjusts the gradients to improve robustness, but this results in limitations in performance. The reconstruction error-based models AE and MemAE with LOE show low robustness due to the high normal loss, as the potential anomalies are treated as normal instances. Meanwhile, DSVDD demonstrates low stability due to its low normal loss. IAD improves robustness by using relative weights, but it does not fully exclude outliers during training. Although LOE and IAD achieve higher robustness than MSE, AAR delivers the most balanced results across all datasets by 0.014-0.052. Furthermore, AAR outperforms the robust loss functions when the dataset is contaminated, demonstrating that robust performance can be achieved without relying on complex loss functions, such as those that combine normal and abnormal losses or use relative weights.

\begin{table}[t]
\small
\caption{AUROC of robust loss functions on image datasets.}
\label{tab:robustloss}
\begin{center}

\begin{tabular}{cccccccccc}
\hline
Dataset   &Model& $\gamma_0$   & MSE & Huber & LOE & IAD & AAR \\
\hline
\multirow{9}{*}{\rotatebox[]{90}{MNIST}}
& \multirow{3}{*}{AE}

 &0  &0.933    	&0.933	  &\textbf{0.934}    &0.934	&0.929\\
&&0.1&0.866	 &0.859	   &0.874	 &0.871	 &\textbf{0.892}  \\
&&Avg.&0.900	 &0.896	   &0.904	 &0.903	 &\textbf{0.910} \\

\cline{2-8}

& \multirow{3}{*}{MemAE}
& 0 & 0.928 & 0.900 & \textbf{0.933} & 0.932 & 0.923 \\ 
&& 0.1 & 0.851 & 0.807 & 0.861 & 0.862 & \textbf{0.887} \\ 
&& Avg. & 0.889 & 0.853 & 0.897 & 0.897 & \textbf{0.905} \\ 

\cline{2-8}

& \multirow{3}{*}{DSVDD}
& 0 & 0.927 & \textbf{0.928} & 0.846 & 0.925 & 0.911 \\ 
&& 0.1 & 0.846 & 0.837 & 0.869 & 0.859 & \textbf{0.876} \\ 
&& Avg. & 0.887 & 0.882 & 0.857 & 0.892 & \textbf{0.894} \\ 
\hline

\multirow{9}{*}{\rotatebox[]{90}{F-MNIST}}
& \multirow{3}{*}{AE}

& 0 & 0.884 & \textbf{0.890} & 0.885 & 0.885 & 0.887 \\ 
&& 0.1 & 0.827 & 0.832 & 0.834 & 0.834 & \textbf{0.865} \\ 
&& Avg. & 0.855 & 0.861 & 0.860 & 0.860 & \textbf{0.876} \\ 

\cline{2-8}

& \multirow{3}{*}{MemAE}
& 0 & 0.888 & 0.891 & 0.890 & 0.891 & \textbf{0.893} \\ 
&& 0.1 & 0.817 & 0.809 & 0.816 & 0.824 & \textbf{0.863} \\ 
&& Avg. & 0.852 & 0.850 & 0.853 & 0.857 & \textbf{0.878} \\ 

\cline{2-8}

& \multirow{3}{*}{DSVDD}
& 0 & \textbf{0.924} & 0.922 & 0.903 & 0.922 & 0.919 \\ 
&& 0.1 & 0.855 & 0.856 & \textbf{0.910} & 0.877 & 0.908 \\ 
&& Avg. & 0.889 & 0.889 & 0.907 & 0.899 & \textbf{0.913} \\ 
\hline

\end{tabular}
\end{center}
\end{table}

\subsection{Experiments on Tabular Datasets}
Table~\ref{tab:tabular} presents the results of anomaly detection on tabular datasets with 20\% contamination. AAR achieves the highest average AUROC across various models. Notably, AAR outperforms MZ by 0.34 (AE), 0.28 (MemAE), and 0.32 (DSVDD) in average AUROC, demonstrating that soft rejection enhances robustness. While MZ exhibits slightly better performance on a few datasets, the differences are minimal and negligible compared to AAR's significant improvements across other datasets. Table~\ref{tab:tabular_total}
shows the average AUROC of comparison methods on thirty tabular datasets with a 20\% contamination ratio. AAR consistently achieves superior performance compared to others. The results not only highlight AAR's robustness and adaptability in managing varying contamination levels but also emphasize its versatility across both image and tabular datasets.

\begin{table}[t]
\small

\caption{AUROC of MZ and AAR on 20\% contaminated tabular datasets.}
\label{tab:tabular}
\begin{center}
\begin{tabular*}{0.7\linewidth}{@{\extracolsep\fill}ccccccc}
\hline
\multirow{2}{*}{Dataset}   & \multicolumn{2}{c}{AE}   &
\multicolumn{2}{c}{MemAE} & 
\multicolumn{2}{c}{DSVDD} \\
\cmidrule{2-3}
\cmidrule{4-5}
\cmidrule{6-7}
& 
MZ & AAR &
MZ & AAR &
MZ & AAR 
\\
\hline
        wine        & \textbf{0.144} & 0.131            & \textbf{0.153} & 0.139            & 0.423 & \textbf{0.467} \\ 
        lympho      & 0.517 & \textbf{0.582}            & 0.513 & \textbf{0.586}            & 0.657 & \textbf{0.665}\\ 
        glass       & 0.799 & \textbf{0.802}            & \textbf{0.799} & 0.782            & \textbf{0.671} & 0.670 \\ 
        vertebral   & 0.583 & \textbf{0.593}            & 0.565 & \textbf{0.573}            & 0.436 & \textbf{0.444} \\ 
        wbc         & 0.688 & \textbf{0.688}            & \textbf{0.704} & \textbf{0.704}   & 0.863 & \textbf{0.871} \\ 
        ecoli       & 0.582 & \textbf{0.616}            & 0.605 & \textbf{0.632}            & 0.708 & \textbf{0.710} \\ 
        ionosphere  & 0.825 & \textbf{0.827}            & 0.838 & \textbf{0.839}            & \textbf{0.884} & 0.882 \\ 
        arrhythmia  & 0.789 & \textbf{0.794}            & \textbf{0.786} & 0.785            & 0.758 & \textbf{0.781} \\ 
        breastw     & \textbf{0.937} & 0.905            & \textbf{0.930} & 0.904            & 0.979 & \textbf{0.981} \\ 
        pima        & 0.666 & \textbf{0.676}            & 0.668 & \textbf{0.680}            & 0.624 & \textbf{0.636} \\ 
        vowels      & 0.738 & \textbf{0.749}            & 0.727 & \textbf{0.734}            & 0.677 & \textbf{0.679} \\ 
        letter      & 0.682 & \textbf{0.700}            & 0.681 & \textbf{0.695}            & 0.653 & \textbf{0.672} \\ 
        cardio      & 0.841 & \textbf{0.925}            & 0.846 & \textbf{0.924}            & 0.805 & \textbf{0.907} \\ 
        seismic     & \textbf{0.665} & 0.658            & 0.658 & \textbf{0.666}            & \textbf{0.641} & 0.608 \\ 
        musk        & 0.071 & \textbf{0.146}            & 0.054 & \textbf{0.100}            & \textbf{0.479} & 0.397 \\ 
        speech      & 0.397 & \textbf{0.437}            & 0.411 & \textbf{0.424}            & 0.643 & \textbf{0.661} \\ 
        thyroid     & 0.951 & \textbf{0.969}            & 0.949 & \textbf{0.969}            & 0.800 & \textbf{0.843} \\ 
        abalone     & \textbf{0.898} & 0.890            & \textbf{0.896} & 0.893            & 0.869 & \textbf{0.870} \\ 
        optdigits   & \textbf{0.147} & 0.136            & \textbf{0.112} & 0.101            & 0.224 & \textbf{0.363} \\ 
        satimage    & 0.584 & \textbf{0.722}            & 0.534 & \textbf{0.623}            & 0.953 & \textbf{0.981} \\ 
        satellite   & \textbf{0.793} & 0.793            & 0.786 & \textbf{0.793}            & 0.826 & \textbf{0.831} \\ 
        pendigits   & 0.208 & \textbf{0.209}            & \textbf{0.192} & 0.176            & 0.395 & \textbf{0.454} \\ 
        annthyroid  & 0.692 & \textbf{0.698}            & \textbf{0.703} & 0.697            & 0.634 & \textbf{0.641} \\ 
        mnist-tabular   & 0.630 & \textbf{0.882}            & 0.650 & \textbf{0.854}            & 0.614 & \textbf{0.662} \\ 
        mammography & \textbf{0.621} & 0.596            & 0.561 & \textbf{0.588}            & 0.704 & \textbf{0.722} \\ 
        shuttle     & \textbf{0.911} & \textbf{0.911}   & 0.877 & \textbf{0.906}            & 0.978 & \textbf{0.981} \\ 
        kdd-rev     & 0.309 & \textbf{0.400}            & 0.356 & \textbf{0.441}            & \textbf{0.764} & 0.763 \\ 
        mulcross    & 0.608 & \textbf{0.609}            & 0.727 & \textbf{0.731}            & 0.514 & \textbf{0.841} \\ 
        forestcover & 0.511 & \textbf{0.744}            & 0.540 & \textbf{0.725}            & 0.427 & \textbf{0.560} \\ 
        kdd         & 0.868 & \textbf{0.872}            & \textbf{0.824} & 0.812            & 0.807 & \textbf{0.836} \\ 
        
        \hline
        
        Avg. & 0.622 & \textbf{0.655} & 0.621 & \textbf{0.649} & 0.680 & \textbf{0.713} \\ 
        \hline
\end{tabular*}
\end{center}
\end{table}

\begin{table}
    \centering    
    \caption{Average AUROC on 20\% contaminated tabular datasets.}
    \label{tab:tabular_total}
    \begin{tabular}{ccccc}
    \hline
        \multicolumn{2}{c}{Methods}&AE & MemAE & DSVDD \\ \hline
        
        \multirow{4}{*}{\rotatebox[origin=c]{90}{\makecell{Robust\\Loss}}}
        &MSE & 0.562	&0.562	&0.578\\
        &Huber & 0.579	&0.578	&0.576\\
        &LOE & 0.629	&0.622	&0.686\\
        &IAD & 0.607	&0.598	&0.629\\
        
        \hline
        
        \multirow{6}{*}{\rotatebox[origin=c]{90}{\makecell{Sample\\Selection}}}
        &10\% Rej. & 0.624	&0.618	&0.647 \\
        &IQR & 0.618	&0.616	&0.676 \\
        &MZ & 0.622	&0.621	&0.680\\
        &QMCD & 0.614	&0.608	&0.656\\
        &$\gamma$GMM &0.651	&0.645	&0.672 \\
        &AAR & \textbf{0.655}	&\textbf{0.649}	&\textbf{0.713}\\
        \hline
    \end{tabular}
\end{table}

\subsection{Experiments across Contamination Ratio} 
\begin{figure*}
  \centering
  \includegraphics[width=0.97\textwidth]{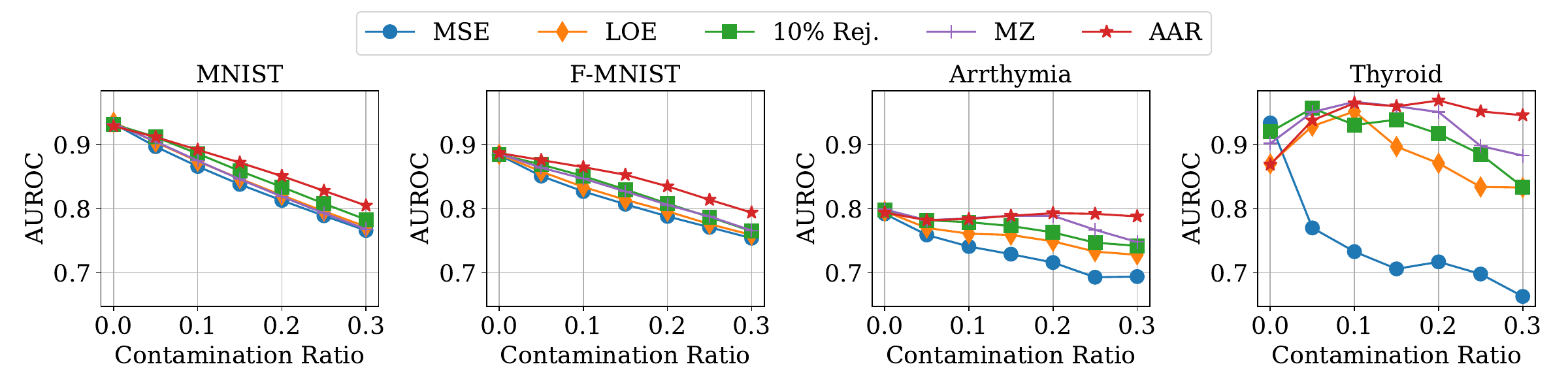}
  \caption{Evaluation across various contamination ratios by training the AE model on MNIST, F-MNIST, arrhythmia, and thyroid datasets, incorporating representative robust methods for comparison.}
  \label{fig:3}
\end{figure*}

We evaluated our methods with AE at various contamination ratios in Figure~\ref{fig:3}. The results reveal that the fixed rejection ratio suppresses the performance of LOE and 10\% rejection, whereas MZ outperforms them through adaptive rejection. However, AAR consistently achieves the best performance across all contamination ratios on image datasets with aggressive rejection. Furthermore, the gap between AAR and comparison methods increases as the contamination ratio increases on both image and tabular datasets. The experiments conclude that aggressive rejection is effective across various contamination ratios, highlighting the importance of aggressive rejection for achieving robustness.

\subsection{Sensitivity Study}

\begin{figure}
  \centering
  \includegraphics[width=0.7\linewidth]{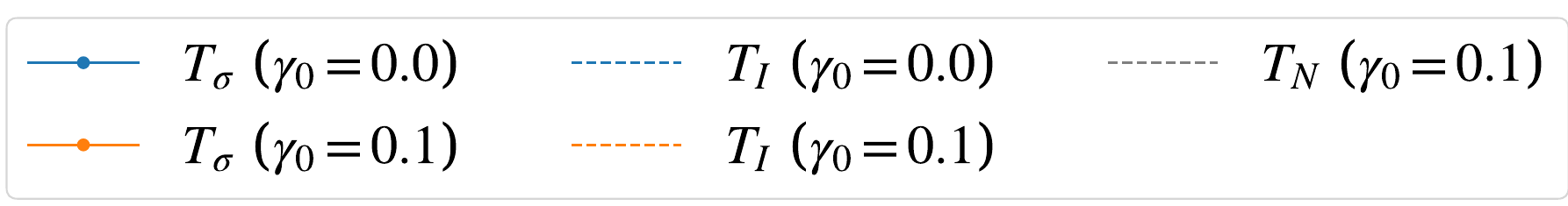}
  \hfill
  \includegraphics[width=0.40\linewidth]{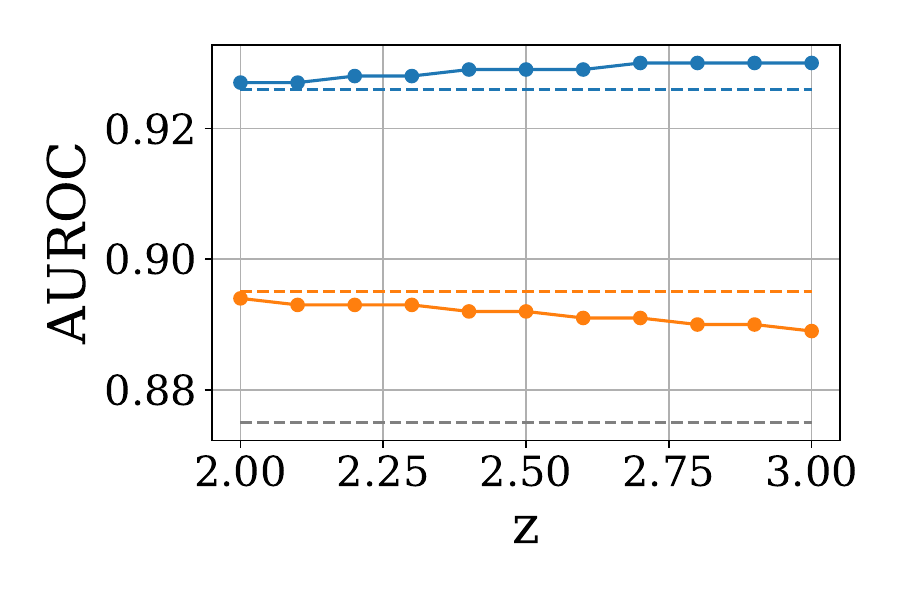}
  \includegraphics[width=0.40\linewidth]{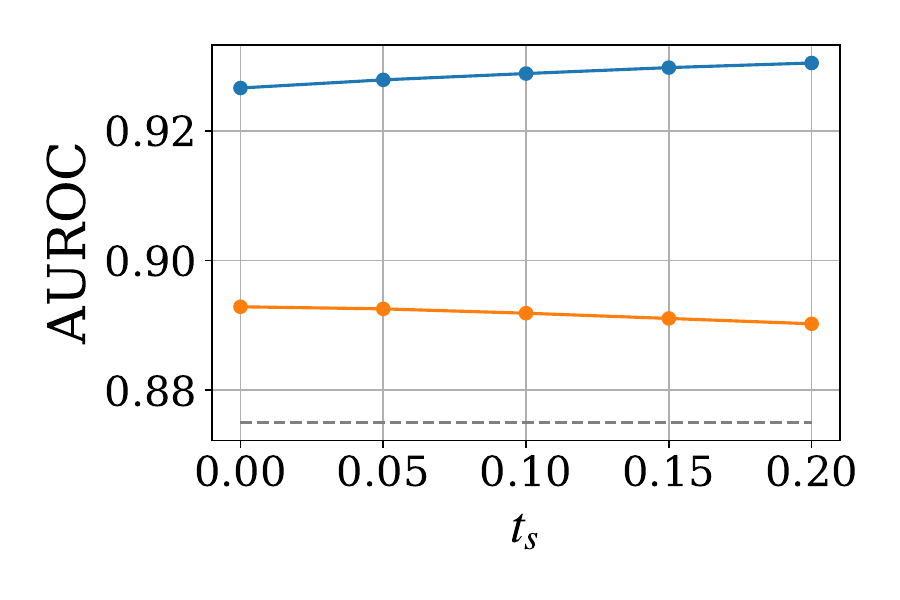}

   \caption{Sensitivity of $z$ (left) and $t_s$ (right) on MNIST dataset with AE. Bold, colored dashed line, and grey dashed line indicates AUROC using $\tau_{\sigma}$ which depends on $z$ value, intersection threshold $\tau_{I}$, and only MZ threshold, respectively.}
   \label{fig:4}
\end{figure}

To address the stability limitations of the intersection threshold ($\tau_{I}$) on clean datasets, we propose the use of $\tau_{\sigma}$. Figure~\ref{fig:4} illustrates the performance of the AE model on both clean and contaminated MNIST datasets as $z$ and $t_s$ vary. The left figure shows that a low $z$ value results in performance closer to that of $\tau_{I}$, while higher $z$ values improve stability and maintain robustness comparable to $\tau_{I}$ and superior to MZ. The right figure demonstrates that incorporating soft rejection enhances model stability while preserving robustness. Specifically, when $t_s=0.1$, stability improves by 0.002 compared to $t_s=0$, while robustness remains higher than that achieved by MZ. These results validate that the $z$-$\sigma$ threshold, combined with soft rejection, provides superior stability compared to $\tau_{I}$ and effectively maintains greater robustness than MZ.

%-------------------------------------------------------------------------
\section{Conclusions}

We propose Adaptive and Aggressive Rejection (AAR), a novel 
anomaly detection framework that dynamically evolves its rejection policy as the model learns.
In the early phase, AAR applies a robust hard rejection -- anchoring training on high-confidence normal samples via a modified z-score threshold -- to stabilize optimization.
As anomaly scoring matures, it transitions to aggressive rejection, deliberately setting a higher cutoff than the true contamination ratio to maximize robustness when the normal and abnormal score distributions overlap.
At this stage, AAR fits a Gaussian Mixture Model to estimate the normal-anomaly boundary to down-weighting borderline samples instead of discarding them outright.
Through theoretical analysis and experiments on 20$\%$ contaminated data, we demonstrate that this multi-stage, distribution-aware scheme yields a 0.041 AUROC improvement over the state-of-the-art, validating AAR's ability to reject ambiguous anomalies more effectively without sacrificing stability. 
Future work will investigate dynamic tuning of rejection weights and z-score parameters in response to real-time contamination estimates, and explore semi-supervised extensions to better handle long-tailed or non-Gaussian score distributions.
We also plan to deploy AAR in critical domains such as medical imaging, financial fraud monitoring, and cybersecurity to assess its real-world impact.
In conclusion, by combining conservative early stage filtering with an aggressive, model-driven thresholding strategy, AAR delivers a scalable, efficient, and balanced solution for anomaly detection under varying contamination conditions.

\appendix
\section{Implementation Details}
\subsection{Image Dataset}
The batch size and number of epochs are set to 256 and 100, respectively, except for DSVDD, which uses 150 epochs for pre-training and 100 epochs for the remainder of the training. The parameters are updated using the Adam optimizer~\cite{kingma2014adam} with a learning rate of 0.0001 and a weight decay of $10^{-6}$. As previously mentioned, the hyperparameters $E$, $z$, and $t_s$ are set to 15, 2.5, and 0.1, respectively.

The architectures of AE and MemAE are based on the design outlined in Gong~\textit{et al.} ~\cite{gong2019memorizing}. On the MNIST and FashionMNIST datasets, the encoder consists of three convolutional modules, each comprising convolution, batch normalization~\cite{ioffe2015batch}, and leaky ReLU activation~\cite{xu2015empirical}, with 16, 32, and 64 filters, respectively. The kernel and stride sizes are both set to 3 and 2. For DSVDD, the autoencoder architecture follows the design described in Ruff~\textit{et al.}~\cite{ruff2018deep}. The encoder consists of two convolutions with 8$\times$5$\times$5 filters and 4$\times$5$\times$5 filters, followed by a final fully connected layer with 32 units. Batch normalization, leaky ReLU, and (2$\times$2)-max-pooling are applied after the convolutions. The biases in the layers of DSVDD are removed to prevent trivial solutions, as reported in Ruff~\textit{et al.}~\cite{ruff2018deep}. The decoder is symmetric to the encoder, with convolutions replaced by deconvolutions and max-pooling replaced by up-sampling. The last deconvolution layer does not include any additional operations, such as batch normalization.

\subsection{Tabular Dataset}
A train-to-test ratio of 1:1 is used~\cite{shenkar2021anomaly}, with contaminated data generated by adding zero-mean Gaussian noise to the actual anomalies~\cite{qiu2022latent}. Preprocessing involves standardization followed by min-max scaling~\cite{ruff2019deep}. The training setup for tabular datasets mirrors that of image datasets, except for batch configurations. Table~\ref{tab:dataset_summary} provides detailed information on dataset composition, batch sizes, and neural network architectures. Batch sizes are adjusted based on the number of training samples and are set to 4096, 1024, 512, 128, or 32. The network architecture consists of fully connected layers, followed by batch normalization and leaky ReLU, with the hidden layer sizes determined by the dimensionality of the tabular data.

\begin{table}[th]
\footnotesize
\centering
\caption{Summary of datasets with size (n), dimension (d), outlier statistics, batch size (b) and hidden layers (h).}
\begin{tabular}{cccccc}
\hline
Dataset & n & d & Outlier & b & h \\ 
\hline
wine             & 129        & 13         & 10 (7.7\%)        & 32     & [32, 16, 8]\\ 
lympho           & 148        & 18         & 6 (4.1\%)         & 32     & [32, 16, 8]\\ 
glass            & 214        & 9          & 9 (4.2\%)         & 32     & [32, 16, 8]\\  
vertebral        & 240        & 6          & 30 (12.5\%)       & 32     & [32, 16, 4]\\ 
wbc              & 278        & 30         & 21 (5.6\%)        & 32     & [32, 16, 8]\\ 
ecoli            & 336        & 7          & 9 (2.6\%)         & 32     & [32, 16, 4]\\ 
ionosphere       & 351        & 33         & 126 (36\%)        & 32     & [32, 16, 8]\\ 
arrhythmia       & 452        & 274        & 66 (15\%)         & 32     & [128, 64, 32]\\ 
breastW          & 683        & 9          & 239 (35\%)        & 32     & [32, 16, 8]\\ 
pima             & 768        & 8          & 268 (35\%)        & 32     & [32, 16, 4]\\ 
vowels           & 1456       & 12         & 50 (3.4\%)        & 128    & [32, 16, 8]\\ 
letter           & 1600       & 32         & 100 (6.25\%)      & 128    & [32, 16, 8]\\ 
cardio           & 1831       & 21         & 176 (9.6\%)       & 128    & [32, 16, 8]\\ 
seismic          & 2584       & 11         & 170 (6.5\%)       & 128    & [32, 16, 8]\\ 
musk             & 3062       & 166        & 97 (3.2\%)        & 128     & [128, 64, 32]\\  
speech           & 3686       & 400        & 61 (1.65\%)       & 512    & [128, 64, 32] \\ 
thyroid          & 3772       & 6          & 93 (2.5\%)        & 512    & [32, 16, 4] \\ 
abalone          & 4177       & 9          & 29 (0.69\%)       & 128    & [32, 16, 4] \\ 
optdigits        & 5216       & 64         & 150 (3\%)         & 512    & [32, 16, 8] \\ 
satimage         & 5803       & 36         & 71 (1.2\%)        & 512    & [32, 16, 8] \\ 
satellite        & 6435       & 36         & 2036 (32\%)       & 512    & [32, 16, 8] \\ 
pendigits        & 6870       & 16         & 156 (2.27\%)      & 1024   & [32, 16, 8] \\ 
annthyroid       & 7200       & 6          & 534 (7.42\%)      & 1024   & [32, 16, 4]  \\ 
mnist-tabular       & 7603       & 100        & 700 (9.2\%)       & 1024    & [64, 32, 16]\\  
mammography      & 11183      & 6          & 260 (2.32\%)      & 1024     & [32, 16, 4]\\  
shuttle          & 49097      & 9          & 3511 (7\%)        & 4096     & [32, 16, 8]\\  
kdd-rev          & 121597     & 120        & 24319 (20\%)      & 4096     & [64, 32, 16]\\  
mulcross         & 262144     & 4          & 26214 (10\%)      & 4096     & [32, 16, 4]\\  
forestcover      & 286048     & 10         & 2747 (0.9\%)      & 4096   & [32, 16, 8]  \\ 
kdd              & 494021     & 120        & 97277 (19.6\%)    & 4096   & [64, 32, 16]\\ 
\hline
\end{tabular}

\label{tab:dataset_summary}
\end{table}

% \bibliographystyle{unsrt}  
% \bibliography{references}

\end{document}